\documentclass[lettersize,journal]{IEEEtran}
\usepackage{graphicx}
\usepackage{graphics}
\usepackage{subfigure}
\usepackage{amsmath,amsfonts,amsthm}
\usepackage{algorithmic}
\usepackage{algorithm}
\usepackage{array}
\usepackage{textcomp}
\usepackage{stfloats}
\usepackage{url}
\usepackage{verbatim}
\usepackage{color}
\usepackage{xcolor}
\usepackage{cite}
\usepackage{makecell}

\begin{document}

\title{Implementation of Big AI Models for Wireless Networks with Collaborative Edge Computing}

\author{Liekang Zeng,
Shengyuan Ye,
Xu Chen,
and Yang Yang
\thanks{Contact: Liekang Zeng (zenglk3@mail2.sysu.edu.cn).}
}

\maketitle

\begin{abstract}
Big Artificial Intelligence (AI) models have emerged as a crucial element in various intelligent applications at the edge, such as voice assistants in smart homes and autonomous robotics in smart factories.
Training big AI models, e.g., for personalized fine-tuning and continual model refinement, poses significant challenges to edge devices due to the inherent conflict between limited computing resources and intensive workload associated with training.
Despite the constraints of on-device training, traditional approaches usually resort to aggregating training data and sending it to a remote cloud for centralized training. Nevertheless, this approach is neither sustainable, which strains long-range backhaul transmission and energy-consuming datacenters, nor safely private, which shares users' raw data with remote infrastructures.
To address these challenges, we alternatively observe that prevalent edge environments usually contain a diverse collection of trusted edge devices with untapped idle resources, which can be leveraged for edge training acceleration.
Motivated by this, in this article, we propose collaborative edge training, a novel training mechanism that orchestrates a group of trusted edge devices as a resource pool for expedited, sustainable big AI model training at the edge.
As an initial step, we present a comprehensive framework for building collaborative edge training systems and analyze in-depth its merits and sustainable scheduling choices following its workflow.
To further investigate the impact of its parallelism design, we empirically study a case of four typical parallelisms from the perspective of energy demand with realistic testbeds.
Finally, we discuss open challenges for sustainable collaborative edge training to point to future directions of edge-centric big AI model training.
\end{abstract}


\section{Introduction}

Big Artificial Intelligence (AI) models are making transformative and disruptive impacts in human-centric intelligent services. 
Their ability to learn, analyze, and process vast amounts of data enables them to perform advanced tasks such as continuous multi-round dialogue, generative artistic creation, and high-precision pattern recognition.
Driven by them, the network infrastructure's edge and wireless networks have witnessed a rapidly growing number of big AI model based applications deployed around users \cite{xu2024unleashing}, e.g., voice-controlled assistants in smart homes and autonomous robotics in smart factories, etc.

Given the human-in-the-loop nature of versatile edge services
\cite{zhou2019edge}, realizing sustainable model refinement and privacy-preserving personalization is of urgent emphasis to fully unleash the advanced ability of big AI models.
Nevertheless, big AI models, which are elfevident to be large, typically comprised of computation-intensive Transformer blocks in millions and even billions of parameters, posing significant challenges for model training on edge devices.
Towards that, traditional approaches usually appeal to the powerful cloud, applying a centralized training mechanism that collects data from all edge devices and distributes models back after training completion.
Although such a mechanism enjoys training acceleration through remote computing resource access, its dependence on the cloud may incur tremendous carbon tax and inevitably raise privacy issues.
To eliminate this cloud dependence, some researchers explore training AI models in situ, which completely reserves model training on the device locally.
While it can well satisfy the privacy demand, the resource-hungry training workload of big AI models makes resource-poor edge devices unaffordable and unsustainable to finish an expected training task.
For instance, training one iteration of the GPT-2 model with a batch size of 128 and a sequence length of 32 requires at least 4.2GB of memory and 2.8TFLOPs. In contrast, typical edge devices in wireless networks usually are equipped with 4GB or 8GB RAM and merely a single mobile SoC.
In summary, existing mechanisms fall short of simultaneously embracing performance, sustainability, and privacy for edge model training.

\begin{figure}[t]
    \centering
    \includegraphics[width=0.9\linewidth]{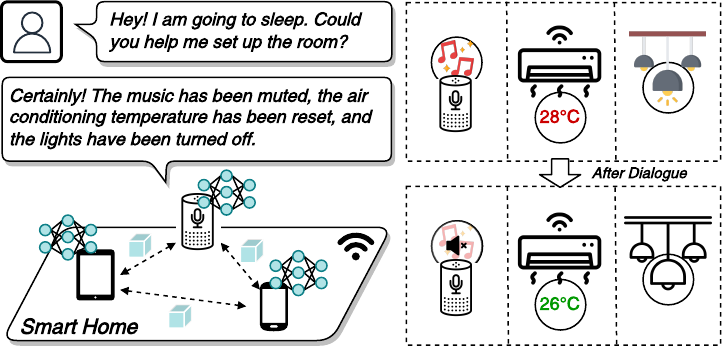}
    \caption{
    An example scenario of intelligent voice assistant in a smart home, which is driven by a collaboratively trained big AI model.
    }
    \label{fig:application}
\end{figure}

Alternatively, we observe that ubiquitous wireless networks such as smart homes and smart factories often consist of a diverse range of reliable dormant devices managed by the same user or organization and in close physical proximity \cite{yang20226g}.
This serves as the impetus to consider nearby accessible edge devices within a trusted domain as a resource pool and engage in collaborative efforts with them in a distributed manner, aiming to expedite private big AI model training at the edge.
As an example, Fig. \ref{fig:application} shows how a voice-controlled smart speaker serves the user's query, where its core AI model is trained by collaborating with available edge devices owned by the same user.
Since all the training courses and data flows are provisioned within the private smart home, privacy issues are completely reserved as those in on-device training. 

Motivated by the above observation, in this article, we propose a novel edge model training mechanism called \textbf{collaborative edge training}, which breaks the resource wall across distributed trusted edge devices in wireless networks for sustainable, expedited, and private big AI model training (e.g., personalized fine-tuning and continual refined learning).
Collaborative edge training borrows the idea of collaborative edge computing \cite{ning2018green, huang2021collaborative, cai2023collaboration} but develops a more concrete mechanism tailored to big AI models.
It targets particularly big AI models and proposes a comprehensive framework to tame the whole lifecycle of edge model training beyond general workload.
Note that collaborative edge training also differs from decentralized learning as it does not necessarily require clients to maintain a complete local model and allows flexible data exchange (raw data, intermediate tensors, etc.) between devices within the trusted domain (while decentralized learning usually only permits transmission of model gradient under privacy restrictions).

In summary, this paper makes the following contributions.
First, to make a thorough investigation on the feasibility of collaborative edge training, we compare it in detail with established methodologies of edge model training (Sec. II).
This comparison includes centralized cloud training, on-device training, and centralized/decentralized federated learning, thereby shedding light on the largely unexplored possibilities of collaboration inherent within wireless edge networks, while simultaneously identifying potential scenarios where they can be utilized effectively.
Next, a comprehensive framework is introduced and sustainable scheduling choices are discussed to facilitate collaborative edge training, encompassing the full life cycle of the training pipeline (Sec. III).
Third, as a case study, a particular focus is placed on investigating the effects of various forms of applied parallelism on training-associated energy consumption, and a thorough analysis is presented, accompanied by performance results derived from realistic testbed experimentation (Sec. IV).
Finally, an extensive discussion is embarked upon, taking a full-stack perspective, to suggest the open challenges pertaining to sustainability in the context of collaborative edge training (Sec. V).

\begin{figure*}[t]
    \centering
    \subfigure[Centralized cloud training.]{
        \begin{minipage}[t]{0.28\linewidth}
        \centering
        \includegraphics[width=\linewidth]{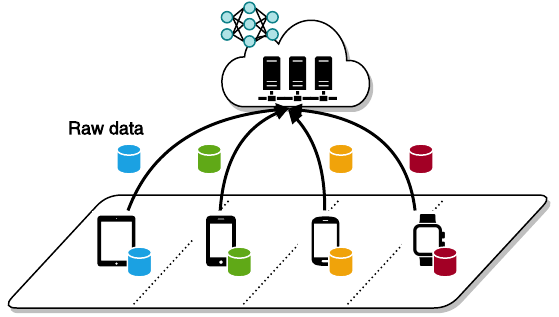}
        \label{fig:training-cloud}
        \end{minipage}
    }
    \
    \subfigure[On-device training.]{
        \begin{minipage}[t]{0.28\linewidth}
        \centering
        \includegraphics[width=\linewidth]{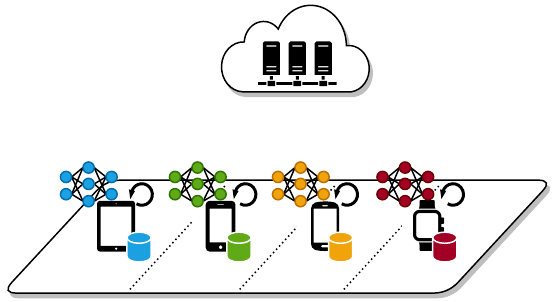}
        \label{fig:training-local}
        \end{minipage}
    }
    \\
    \subfigure[Centralized federated learning.]{
        \begin{minipage}[t]{0.28\linewidth}
        \centering
        \includegraphics[width=\linewidth]{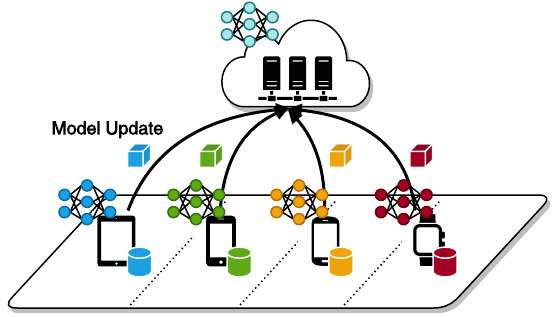}
        \label{fig:training-central-fl}
        \end{minipage}
    }
    \
    \subfigure[Decentralized federated learning.]{
        \begin{minipage}[t]{0.28\linewidth}
        \centering
        \includegraphics[width=\linewidth]{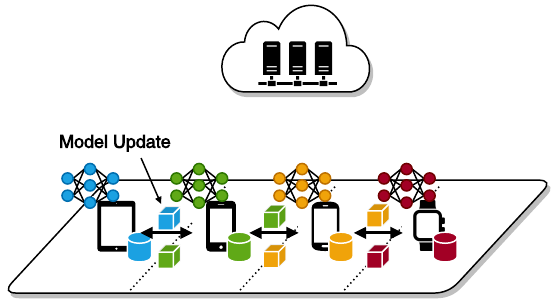}
        \label{fig:training-decentral-fl}
        \end{minipage}
    }
    \
    \subfigure[Collaborative edge training.]{
        \begin{minipage}[t]{0.28\linewidth}
        \centering
        \includegraphics[width=\linewidth]{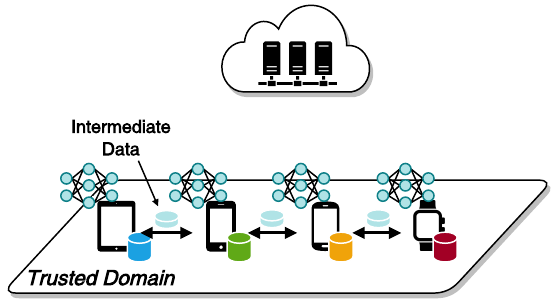}
        \label{fig:training-collaboration}
        \end{minipage}
    }
    \caption{Existing AI model training mechanisms versus collaborative edge training.}
    \label{fig:training}
\end{figure*}

\section{Background and Motivation}

\subsection{Big AI Models and Transformer}

Big AI models have demonstrated exceptional performance in a wide range of tasks, including natural language questions and answers, fine-grained computer vision, and autonomous robotics control.
These complex intelligent models implicate their advanced ability in an extensive number of its parameters, often reaching millions or even billions.
Basically, in many representative big AI models like Bert and GPT, the tremendous parameters are organized in multiple stacked Transformer blocks.

The primary components of a typical Transformer block are the Multi-Head Attention module and Multi-Layer Perceptron (MLP).
These components are interconnected through element-wise operations, including Dropout, Residual Addition, and Layer Norm.
Within the Multi-head Attention block, the initial linear layer generates separate matrices for Query (Q), Key (K), and Value (V) for each attention head.
Self-attention is then independently performed by each head, with the resultant outputs concatenated and subsequently processed through a final linear layer to obtain the overall output.
On the other hand, the MLP module comprises two linear operations that expand the hidden size from \textit{h} to \textit{4h} and subsequently reduce it back to \textit{h}.

A Transformer block can introduce massive matrix multiplications and add operations.
Given the tens or even hundreds of Transformer blocks stacked in a big AI model as well as the considerable training iterations on giant datasets, their training necessarily requires substantial computational resources and ineluctably raises sustainability challenges.

\subsection{Issues of Existing Edge Training Mechanisms}
\label{sec:existing-para}

The development of big AI models continues to advance the capabilities of AI systems and drive progress in various fields, especially for intelligent services at the edge.
To train these big models in wireless edge networks, traditional methods typically resort to centralized cloud training, on-device training, and Federated Learning, as illustrated in Fig. \ref{fig:training}.

\textbf{Centralized cloud training} typically collects raw labeled data from distributed edge clients and employs a remote cloud as the dedicated training platform (Fig. \ref{fig:training-cloud}).
The training workload is entirely reserved and performed on dedicated cloud servers or clusters, which have abundant computational resources and storage capacity.
Albeit it can effectively expedite the training process, the utilization of the remote datacenter comes at a price of excessive carbon footprint, and sharing sensitive data directly across clients and remote datacenters inevitably raises users' private concerns.

\textbf{On-device training} brings the training process closer to the end-user by leveraging the computational power and resources available on their own devices (Fig. \ref{fig:training-local}).
Contrary to the cloud-based approach,  sensitive data does not need to leave the user's device in on-device training, and thus privacy preservation is guaranteed.
It also reduces reliance on network connectivity and cloud infrastructures, ensuring that training can be performed even in offline or low-connectivity scenarios.
Nevertheless, limited computational resources and memory on edge devices restrict the size and complexity of models and make them unsustainable for big AI models.
The isolation of training data on individual devices also restricts the representation ability of models owned by them, since in-situ models can only learn from local data.

\textbf{Centralized federated learning} is a distributed machine learning approach where the training of a model occurs on various edge devices or servers (Fig. \ref{fig:training-central-fl}), instead of solely centralizing the training process on a single server or in the cloud. In federated learning, the training data remains on the individual devices, and only the model updates or aggregated weights are exchanged between the devices and the central server.
However, it still requires each edge device to provision a local model, suffering from the same resource constraints in on-device training.

\textbf{Decentralized federated learning} performs federated learning in a central-server-less manner.
Unlike centralized federated learning that resorts to a central coordinator (e.g., the cloud) to orchestrate the pipeline and train a global model, decentralized federated learning allows clients to exchange model updates directly with each other \cite{beltran2023decentralized}, as in Fig. \ref{fig:training-decentral-fl}.
Nonetheless, it endures the same issue as that in centralized federated learning, where every client bears a complete local model with limited on-device resources.

\subsection{Motivation of Collaborating Edge Training}
\label{sec:motivation}

The above discussion reveals the practical issues of existing training approaches with respect to performance, sustainability, and privacy. 
Alternatively, we observe that many edge scenarios typically contain a set of available (idle) edge devices within a trusted domain and they can be properly managed to render computational resources as a whole for big AI model training.
Motivated by this, we propose collaborative edge training, a distributed training mechanism that orchestrates multiple available edge devices within the trusted domain as a resource pool to provide sustainable edge model training in wireless networks (Fig. \ref{fig:training-collaboration}).

\textbf{Merits.}
Collaborative edge training viably exploits the inherent potential of wireless networks and offers merits on five levels.
First, through harvesting idle computational resources from underutilized devices, collaborative edge training implements resource efficiency, i.e., increased edge resource utilization, compared to on-device training.
Second, by breaking the boundary of training data in the trusted domain, collaborative training unlocks a larger volume of learning materials for training, in contrast to on-device training which is only accessible to local data. 
Third, by exploiting resources within the edge scenario, collaborative edge training fully inherits the benefit of edge computing and profitably reduces server rental in cloud training, i.e., rendering cost efficiency.
Fourth, with an augmented number of edge devices, collaborative edge training enables more fault tolerance opportunities and offers more training robustness than single-device in-situ training.
Fifth, by managing all data flow and computation workload within the trusted domain, collaborative edge training is independent of remote cloud and promises reliable privacy preservation.

\textbf{Distinction from existing approaches.}
Collaborative edge training differs from the existing training approaches discussed in Sec. \ref{sec:existing-para}.
First, compared to centralized federated learning and cloud training, it does not rely on the cloud server, effectively mitigating privacy leaky risks brought by the cloud providers.
Second, unlike on-device training that affords workloads solely on a single device, collaborative edge training assembles multiple facilities for a shared workload, which breaks the resource wall among devices.
Third, different from centralized/decentralized federated learning that enforces each client to hold a full local model, collaborative edge training allows partial model replication on devices and thus enables the system to support big AI models on a larger scale.
Besides, due to the privacy restriction, clients in decentralized federated learning can not communicate raw training data but only model updates; in collaborative edge training, clients instead can freely exchange data (training data, intermediate tensors, etc.) within the trusted domain.
    
\begin{figure*}[t]
    \centering
    \includegraphics[width=0.95\linewidth]{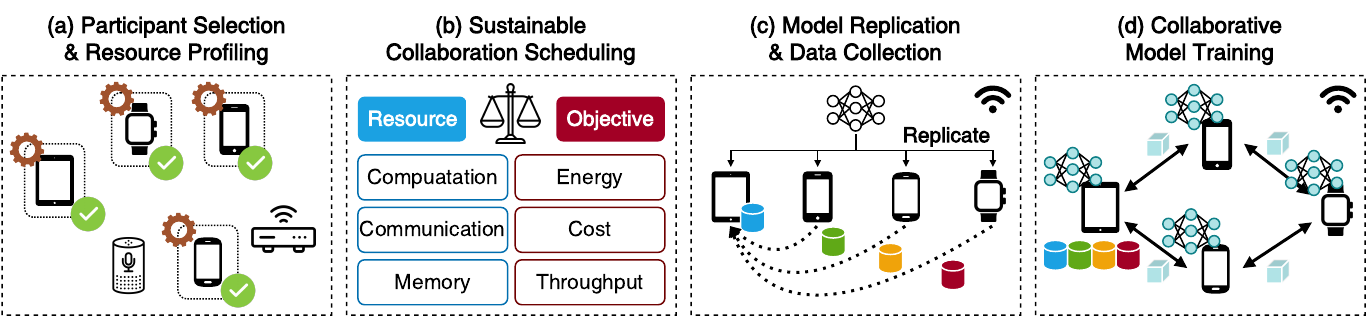}
    \caption{Overview of collaborative edge training workflow.}
    \label{fig:overview}
\end{figure*}

\textbf{Potential scenarios.}
Collaborative edge training is particularly useful for sustainable big model fine-tuning and privacy-preserving personalization in trusted edge domains.
Note that trusted domains are configurable and adjustable, which can be flexibly defined according to the requirements of users' privacy preferences.
For instance, in a smart factory that deploys a cluster of sensors, monitors, and edge servers, these edge devices are managed by the same organization and can be naturally viewed in the same trusted domain \cite{ye2024asteroid}.
In smart homes, family members usually possess several smartphones, smartwatches, tablets, and laptops, and one or multiple separate trusted domain(s) can be established among these devices as long as their owners are willing to share resources \cite{zeng2020coedge}.

\section{Collaborative Edge Training of Big AI Models}

To translate the merits discussed above into practical utilities in deployment, we design a general and compatible framework to unify collaborative edge training in various scenarios.
In what follows, we will present its overview and discuss scheduling choices toward sustainability.

\subsection{Framework Overview}

Fig. \ref{fig:overview} illustrates a comprehensive framework for collaborative edge training in trusted domains, which works in four phases.
Specifically, given a set of trusted edge devices, the system first selects some from them by examining their access availability, computational capability, power requirements, etc.
This is accomplished by resource profiling techniques, e.g., applying benchmark toolkits upon edge facilities, to obtain their performance metadata (Fig. \ref{fig:overview}(a)). 
Next, in the second phase, the system scheduler is committed to determining an orchestration strategy such that available computation, communication, and memory resources are efficiently utilized to realize dedicated sustainability objectives like energy consumption, cost budget, and training throughput (Fig. \ref{fig:overview}(b)).
The orchestration strategy comprises a group of system configurations such as parallelism principles and workload distribution, and will be detailed in the next subsection.
In light of this strategy, the system will replicate the targeted big AI model partially or completely into participated edge devices in the third phase, and direct training data collection to the entrance of the devices' pipeline (Fig. \ref{fig:overview}(c)).
Upon data collected and the model installed, the system will launch model training runtime to complete targeted training tasks,
(Fig. \ref{fig:overview}(d)), exchanging data across wirelessly connected edge devices if necessary.

\subsection{Scheduling Collaborative Edge Training for Sustainability}
\label{sec:scheduling_choics}

Collaborative edge training of big AI models over wireless networks comes with several unique properties.
Specifically, it introduces versatile communicated content, including model parameters, raw training data, intermediate tensors, etc., in different periods of the workflow.
This not only magnifies the communication complexity but also induces large transferred data volume.
Besides, to resolve the frequent data exchange between devices in collaboration, the wireless communications are expected to be accurate and real-time.

The proposed collaborative edge learning framework is compatible with different designs of wireless networks as long as the participated edge devices can be well connected for collaboration.
Despite that, provisioning a collaborative edge training pipeline for sustainable big AI model training is yet up against a set of Research Questions (RQs) to be answered.
Following the four phases in the framework presented above, we discuss four fundamental RQs as follows.

\textit{RQ1: How to select proper participants?}
Selecting proper participants for collaborative edge training is a crucial step to ensure the effectiveness and efficiency of the training process and is the core RQ in the first phase (Fig. \ref{fig:overview}(a)).  
The available devices can be heterogeneous with different types of devices, operating systems, or hardware configurations.
To perform favorable selection among them, several factors should be considered including device capability, data relevance, network connectivity, privacy and security requirements, as well as participation incentives.
Besides, the selection process may very depending on specific applications and requirements, demanding different performance preferences and distinct efforts in striking balances.

\textit{RQ2: How to design optimal parallelism?}
Given the registered devices, the second phase (Fig. \ref{fig:overview}(b)) is obliged to determine a principle to orchestrate the parallel data flow among them.
This principle is imperative to sustainable collaborative training since it dominates the task partitioning of Transformer blocks, load balancing of distributed devices, as well as synchronization and aggregation strategies \cite{ye2024asteroid}. 
Either of them involves the utilization of computation, communication, and memory in the resource pool, and thus parallelism design plays an essential role in collaborative edge training.
Sec. \ref{sec:case_study} will take this RQ as a case study to further explore the impact of different parallelisms in collaborative edge training.

\begin{figure*}[t]
    \centering
    \includegraphics[width=\linewidth]{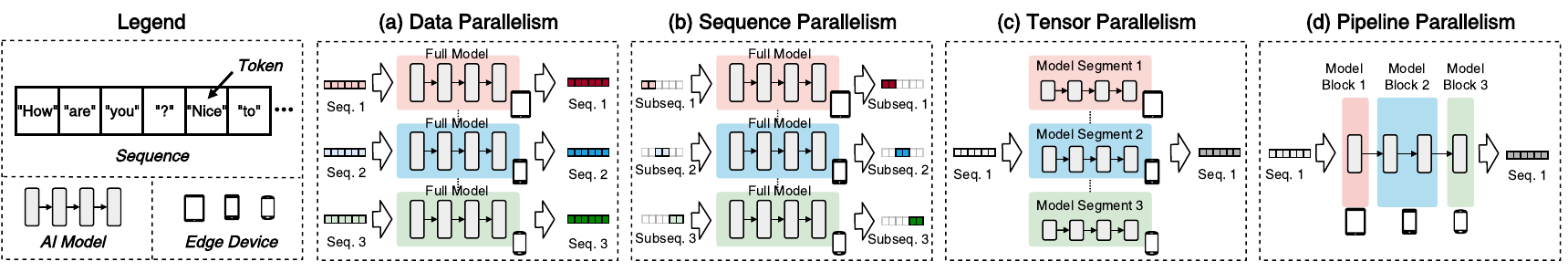}
    \caption{
    Illustration of different parallelisms in collaborative edge training. Different colors represent assignments to different edge devices.
    }
    \label{fig:parallelism}
\end{figure*}

\textit{RQ3: How to arrange device topology?}
Provided a decided parallelism principle, the system in the third phase (Fig. \ref{fig:overview}(c)) is compelled to arrange a topology of the participant devices under the existing connectivity of the wireless network.
In essence, the arrangement of device topology entails finer-grained consideration of training workload distribution, memory footprint of model segments, and transferred data content.
For instance, in pipeline parallel, the big AI model is divided into successive segments and replicated these segments to different devices.
The topology, i.e., the ordering of participated devices, thus impacts which segments the individual devices load, where the raw data is collected, and how the data flows across devices.

\textit{RQ4: How to ensure fault tolerance?}
Maintaining a sustainable training runtime in the fourth phase (Fig. \ref{fig:overview}(d)) needs to guard a continuously stable pipeline throughout the training lifecycle.
Generally, many distributed-computing-oriented fault tolerance techniques are advantageous, such as redundant replication, periodical checkpointing, exception handling, and resilient communication.
Nonetheless, applying them to collaborative edge training demands a careful design with respect to the tradeoff between training performance and recovery overhead.

\section{Case Study: Sustainable Parallelism Design}
\label{sec:case_study}

The four RQs discussed above reflect the scheduling choices in the four phases of collaborative edge training.
This section takes RQ2 as the knob to further investigate how parallelism design impacts system energy demand.
Particularly, we first briefly introduce four typical parallelisms in Transformer-based big AI models, and next empirically evaluate their energy consumption with realistic experiments.

\subsection{Parallelisms for Collaborative Transformers Training}
\label{sec:parallelism}

According to the separation of data, model, and activations, parallel Transformer processing can be categorized into four types: data parallelism, sequence parallelism, tensor parallelism, and pipeline parallelism.
Fig. \ref{fig:parallelism} depicts parallel instances with three trusted devices and a Transformer-based language model, where the input data is a sequence of successive tokens (words, punctuation, etc.) selected from the GLUE dataset \cite{wang2018glue}, as shown in the legend.

\textbf{Data parallelism} is one of the commonly adopted parallelisms that distributes the training data across multiple devices and performs parallel training among them.
As shown in Fig. \ref{fig:parallelism}(a), each device replicates a copy of the full model and is fed with an individual sequence for training.
Every iteration they finish, their models will be reduced via model synchronization, i.e., AllReuce operations, to ensure global consistency, where gradient updates are synchronously exchanged among devices.
The above procedure will repeat until all data batches are processed and all local models are converged.

\textbf{Sequence parallelism} is a customized parallelism for Transformer-based big AI models, which processes input of token sequences.
Fig. \ref{fig:parallelism}(b) illustrates its mechanism, where the input sequence is split into smaller subsequences and each subsequence is independently processed through Transformer blocks in parallel by different devices.
Due to the partitioning in the sequence dimension, sequence parallelism introduces dependencies between the subsequences, as consecutive subsequences may depend on the outputs of previous subsequences.
To address this, intermediate states, such as the attention layer's outputs, are exchanged between devices through the AllGather and AllReduce collective communications to ensure correct calculations. 

\textbf{Tensor Parallelism} focuses on parallelizing the computations in the attention weight dimensions.
Specifically, the self-attention and feed-forward layers in Transformer blocks engage massive matrix multiplications and can be parallelized along the dimension of the head (in Multi-Head Attenion modules) or hidden size (in Multi-Layer Perceptron modules).
Note the input sequence remains as a whole while the embeddings and attention masks are divided into smaller chunks (model segments) and distributed across devices, as in Fig. \ref{fig:parallelism}(c).
To handle data synchronization, tensor parallelism relies on collective operations, such as AllReduce, to exchange gradient information across devices.

\textbf{Pipeline parallelism} partitions the model in the layer dimension, which divides the model into multiple consecutive blocks, and each block is processed by different devices in a pipelined manner (Fig. \ref{fig:parallelism}(d)).
Each device processes its assigned model block independently, taking into account the dependencies between model blocks.
The intermediate outputs of each model block are passed forward to the next model block for further processing.
Pipeline parallelism enables overlapping computations, where one model block can start processing its input while another model block is still processing its previous input.
This feature allows for reducing the idle time of devices and thus increasing the overall throughput of the training process.

Although pipeline parallelism and split learning \cite{vepakomma2018split} both tackle privacy issues and partition neural networks in layer granularity, they are largely different model training mechanisms due to the following reasons.
First, split learning is designed principally for cross-platform training without sharing raw data under a particular premise of privacy, whereas pipeline parallelism in collaborative training allows mutual data sharing across edge devices within the trusted domain, aiming at sustainable and expedited edge learning.
Second, split learning is carried out between two sides (i.e., the client and the server), while pipeline parallelism allows two- or multi-party training over distributed edge devices.
Third, the training data of split learning are wholly from the client side, but pipeline edge training may leverage data from all available participants.

Note that these four parallelisms do not modify the architecture of the AI model and fully reserve all data and model parameters.
All model gradients are synchronously updated.
Hence, their training convergences are guaranteed as it is in single-device training.
Among them, both tensor parallelism and pipeline parallelism enable partial model replication on edge devices, which effectively reduces on-device memory footprint and attains better model scalability.

\begin{figure}[t]
    \centering
    \subfigure[Energy demand in CPU-only mode.]{
        \begin{minipage}[t]{0.45\linewidth}
        \centering
        \includegraphics[width=\linewidth]{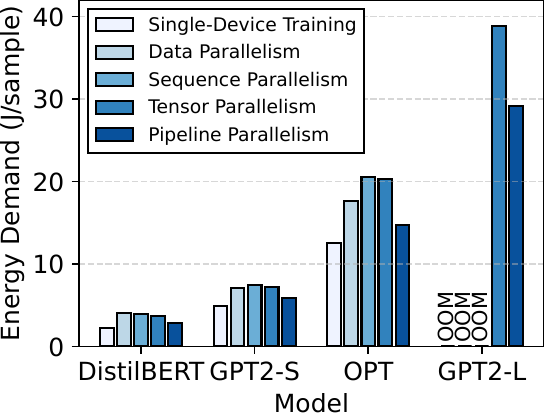}
        \label{fig:energy-homo-cpu}
        \end{minipage}
    }
    \
    \subfigure[Energy demand in GPU-enabled mode.]{
        \begin{minipage}[t]{0.45\linewidth}
        \centering
        \includegraphics[width=\linewidth]{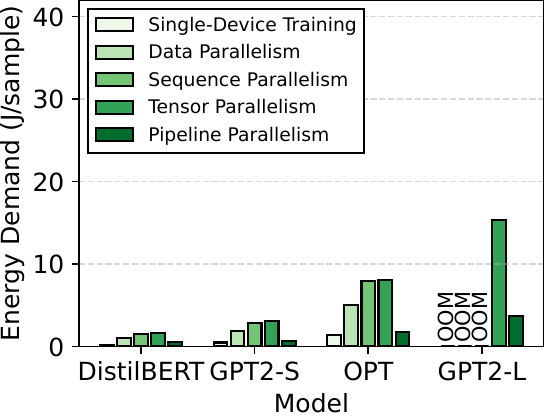}
        \label{fig:energy-homo-gpu}
        \end{minipage}
    }
    \\
    \subfigure[Measured latency in CPU-only mode.]{
        \begin{minipage}[t]{0.45\linewidth}
        \centering
        \includegraphics[width=\linewidth]{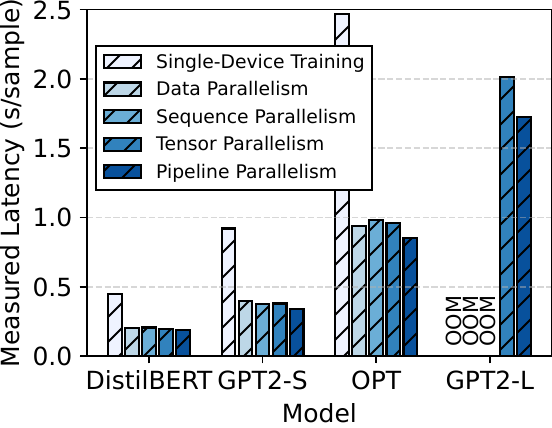}
        \label{fig:latency-homo-cpu}
        \end{minipage}
    }
    \
    \subfigure[Measured latency in GPU-enabled mode.]{
        \begin{minipage}[t]{0.45\linewidth}
        \centering
        \includegraphics[width=\linewidth]{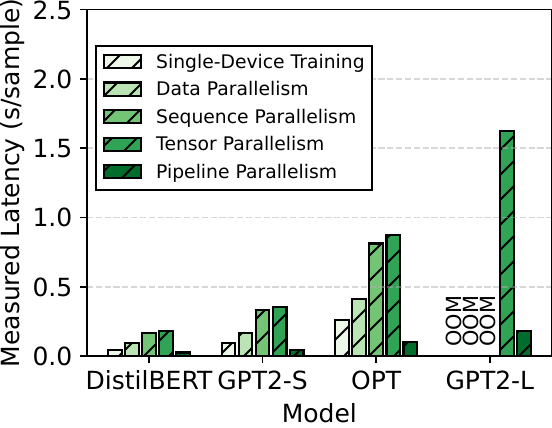}
        \label{fig:latency-homo-gpu}
        \end{minipage}
    }
    \caption{
    Energy demand and measured latency per sample of collaborative edge training under different parallelism in a homogeneous testbed. OOM indicates the out-of-memory error.
    }
    \label{fig:homo}
\end{figure}

\subsection{Empirical Performance Evaluation}

Given the four parallelisms discussed above, we build realistic testbeds to evaluate their sustainability in terms of energy consumption and measured latency empirically. 
In particular, our testbeds comprise two trusted domains, a homogeneous one with four Jetson Nanos and a heterogeneous one with two Jetson Nanos, one Jetson TX2, and one Jeston NX.
Each testbed has two modes, i.e., a CPU-only mode and a GPU-enabled mode, for a thorough evaluation.
The specifications of edge devices are as follows: Jetson Nano has a 1.47GHz Cortex-A53 CPU, a 128-core Maxwell GPU, and 4GB RAM; Jetson TX2 has a 2GHz Cortex-A57 CPU, a 256-core Pascal GPU, and 8GB RAM; Jetson NX has a 6-core 1.4GHz Carmel CPU, a 384-core Volta GPU, and 8GB RAM.
In each domain, we connect all available devices using 1000Mbps wireless networks to emulate the networking environment in a smart home scenario and assume all devices within the trusted domain to participate in the collaboration.
All parallelisms are implemented based on PyTorch, where data parallelism is built using the PyTorch \textit{Distributed Data Parallel} library and performs gradient synchronization every 5 iterations.
To better demonstrate the superiority of collaboration, we include a baseline scenario of single-device training that tests on a Jetson Nano.
Experiments are carried out with four typical big AI models: DistilBERT, GPT2-S, OPT, and GPT2-L, where they are respectively with 6, 12, 24, and 36 Transformer blocks, and 66M, 124M, 350M, and 0.7B parameters.
For model details, they respectively have 12, 12, 16, and 20 attention heads, and 768, 768, 1024, and 1280 hidden sizes.
The training is conducted with a subset of samples, where the average sequence length is 32, from QNLI corpus of the popular GLUE datasets \cite{wang2018glue}.
To ensure a fair comparison, all measurements are in the same global mini-batch size of 128, and are repeated for multiple rounds to export their average.

\begin{figure}[t]
    \centering
    \subfigure[Energy demand in CPU-only mode.]{
        \begin{minipage}[t]{0.45\linewidth}
        \centering
        \includegraphics[width=\linewidth]{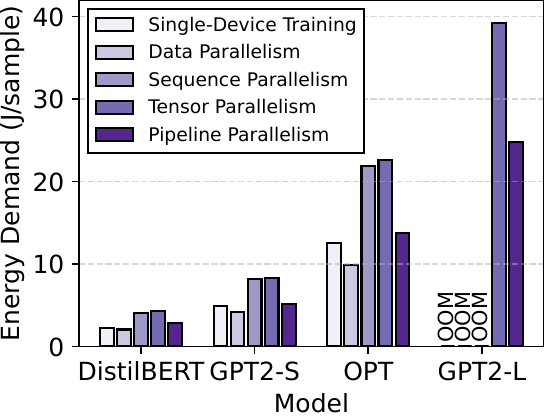}
        \label{fig:energy-hetero-cpu}
        \end{minipage}
    }
    \
    \subfigure[Energy demand in GPU-enabled mode.]{
        \begin{minipage}[t]{0.45\linewidth}
        \centering
        \includegraphics[width=\linewidth]{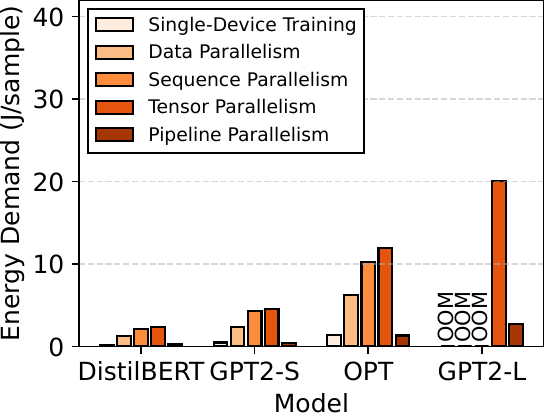}
        \label{fig:energy-hetero-gpu}
        \end{minipage}
    }
    \\
    \subfigure[Measured latency in CPU-only mode.]{
        \begin{minipage}[t]{0.45\linewidth}
        \centering
        \includegraphics[width=\linewidth]{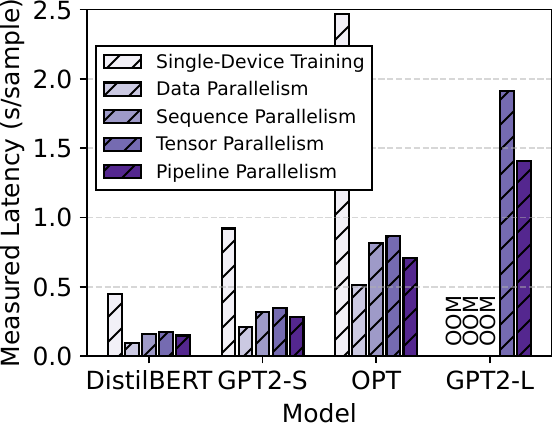}
        \label{fig:latency-hetero-cpu}
        \end{minipage}
    }
    \
    \subfigure[Measured latency in GPU-enabled mode.]{
        \begin{minipage}[t]{0.45\linewidth}
        \centering
        \includegraphics[width=\linewidth]{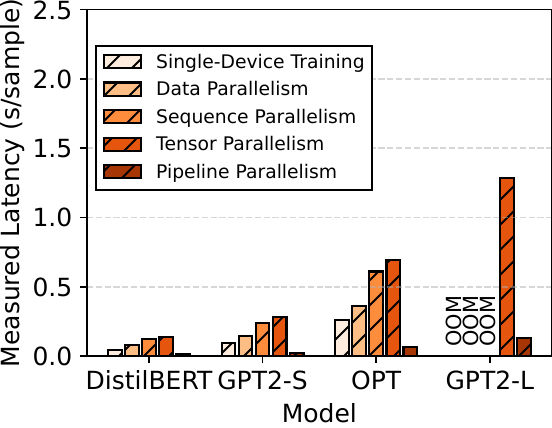}
        \label{fig:latency-hetero-gpu}
        \end{minipage}
    }
    \caption{
    Energy demand and measured latency per sample of collaborative edge training under different parallelism in a heterogeneous testbed. OOM indicates the out-of-memory error.
    }
    \label{fig:hetero}
\end{figure}

We measure system sustainability by the metric of energy demand and measured latency per sample.
Specifically, we launch a training task from scratch and record the total consumed energy/latency and the number of processed data samples within a period, and accordingly calculate sample-wise energy/latency.  
Fig. \ref{fig:homo} and Fig. \ref{fig:hetero} show the experimental results when applying different parallelisms, where their performances vary from models.
From the results, we can derive the following observations.
First, collaborative training is able to render significant training acceleration over single-device training across all models, and by selecting proper parallelism (e.g., pipeline parallelism), it can achieve fair energy consumption on par with single-device training.
In particular for GPT2-L, its parameter scale exceeds the available memory space of a single device, leading single-device training to an out-of-memory error and incomplete training.
Data parallelism and sequence parallelism, which also require to load full models on individual devices, also fail for the same reason.
By contrast, tensor parallelism and pipeline parallelism still finish the training since they allow partial model replication on participating devices, which effectively mitigates the single-device memory usage.
Second, a larger size of parameters leads to higher energy demand, and their growths are positively correlated.
This is particularly evident in homogeneous setups (i.e., Fig. \ref{fig:energy-homo-cpu} and Fig. \ref{fig:energy-homo-gpu}), where OPT with 5.3$\times$ and 2.8$\times$ parameters records 5.1$\times$ and 2.7$\times$ average energy consumption across four parallelisms than DistilBERT and GPT2-S, respectively.
Third, data parallelism and pipeline parallelism perform generally better than sequence parallelism and tensor parallelism.
This is because the latter two principles require much more data synchronization between devices when their sequences or activations are split (c.f. Fig. \ref{fig:parallelism} and Sec. \ref{sec:parallelism}), for which the communication overhead dominates the collaborative training process and exhausts a majority of energy.
Fourth, CPU-only training takes much more energy than GPU-enabled training, no matter for homogeneous or heterogeneous setups.
Transformer block processing is essentially highly parallelized matrix multiplications, which are GPU-friendly operations.
Therefore, compared with CPU-only mode, the GPU-enabled system can availably reduce the processing time of Transformer blocks, as indicated by the latency results, and achieve better energy efficiency.
From the above observations, we can conclude that applying data parallelism and pipeline parallelism with GPU-equipped edge devices can attain superior sustainability among others, while finer-grained hybrid parallelism and multi-tier solutions are desired to exploit further the potential implicated in wireless edge networks.

\section{Open Challenges for Sustainable Collaborative Edge Training}

Despite omnifarious opportunities, collaborative edge training is yet confronted with a set of open challenges toward sustainable and high-performance big AI model training.

\textbf{Sustainability metric design.} 
The foremost question to optimize sustainability is how to define and quantify sustainability.
In general, sustainability pertains to minimizing negative impacts on the environment, society, and resources.
We thus may quantitatively design metrics from ``4E" dimensions: Efficiency, Economics, Environment-friendliness, and Ethics.
Specifically, efficiency means performance efficiency on latency, throughput, accuracy, etc.
Economics tackles financial budgets, expecting a proper price-per-performance such that the AI services can run stably in the market.
Environment-friendliness indicates protecting our environment and considers metrics of carbon emission, resource conservation, and lifecycle hardware assessment \cite{wu2022sustainable}.
Ethics calls for responsible AI, which aligns big AI models with human beings to ensure fairness, transparency, inclusivity, and privacy.

\textbf{Efficient collaboration orchestration.}
Achieving efficient collaboration orchestration is essential because it directly impacts resource utilization across edge devices and thereupon the overall performance and scalability of the collaborative edge training system.
In particular, it involves optimizing resource allocation, task scheduling, and communication strategies to ensure that the training process is carried out efficiently and effectively while utilizing the available edge devices' capabilities to their fullest, especially for supporting big AI models on massive scales.
Besides, fault tolerance is also a key aspect demanding careful design since the wireless edge networks are often dynamic and even vulnerable.
Following the discussion in Sec. \ref{sec:scheduling_choics}, a high-performance system should answer the four RQs through the training lifecycle so as to holistically coordinate and manage the resources and tasks across multiple edge devices involved in the collaborative training process.

\textbf{Participant incentivization.}
Efficient collaborative edge training is indispensable for the participation of multiple edge devices in the trusted domain.
Towards their willingness to share computational, communication, and storage resources, how to incentivize, measure, and pay back their contribution in collaboration can significantly impact the long-term sustainability of the training pipeline.
To address the challenges, researchers and practitioners need to devise effective mechanisms that align the interests of the participants with the collective goals of the collaborative edge training systems.
These mechanisms may include monetary rewards, reputation systems, data and model-sharing agreements, or privacy-preserving techniques.
It is crucial to strike a balance between providing incentives for active participation and ensuring fairness, privacy, and security for all participants.

\textbf{AI-native Wireless networks.}
The design of wireless networks is crucial for the performance of wireless big AI models. 
On the one hand, wireless networks serve as a networking infrastructure to support big AI model computing over edge devices, servers, and the cloud.
The quality of communication can be a bottleneck in rendering sustainable big AI model based services \cite{wu2022revisiting}.
On the other hand, wireless networks themselves can be rich data sources for building and enhancing wireless big AI models.
The design of wireless networks can steer the evolution of wireless big AI models towards better human-centric services.
In the real-world deployment of collaborative edge training, network operators may allocate more bandwidth between participated devices such that their frequent data exchange is carried out with lower delay and higher throughput.
Emerging communication techniques like Ultra-Wideband and 6G \cite{wang2023road} may be useful, but more sustainable, lightweight, and reliable protocols and tools are desired to account for the unique characteristics of wireless edge networks including limited bandwidth, intermittent connectivity, and diverse device capabilities.

\textbf{Wireless-native AI models.}
Traditionally, big AI models are trained and deployed on powerful centralized servers or in the cloud.
However, in collaborative edge training scenarios, the focus shifts towards designing wireless-native models that embrace the wireless networks and the constrained execution environment of edge devices like limited computational capability, memory, and battery life.
To accomplish that, wireless-native AI models may optimize their structure, complexity, and computational requirements to ensure sustainable execution in wireless edge networks.
Several model compacting techniques can be applied to generate these models like neural architecture search, knowledge distillation, and model compression.
For example, one may adopt knowledge transfer techniques to a larger, more complex teacher model from the cloud (e.g., network cloud or base station cloud) and distillate its knowledge to the student model at the edge, so as to improve the training efficiency of edge models.

\textbf{Power-efficient hardware.}
Edge devices, such as IoT sensors and mobile devices, often have limited battery life and energy resources.
Collaborative edge training involves distributed computations across these devices, which can be computationally intensive and drain the battery quickly.
Developing power-efficient hardware solutions is thus crucial to addressing the energy consumption challenge and ensuring sustainable and prolonged operation of edge devices during collaborative training.
Towards this challenge, the community has explored an array of solutions like low-power processors, dynamic power management techniques, energy harvesting devices, and edge AI accelerators.

\textbf{Practical privacy and security.}
A sustainable training system must be trusted and admitted by users such that necessary training data and computational resources are provided.
In the proposed collaborative edge training framework, participating devices are supposed to be in one trusted domain to ensure privacy preservation.
However, this requirement can be properly relaxed if some practical privacy and security techniques are applied and users' privacy is protected at the same level.
For example, one may leverage cryptography techniques, e.g., differential privacy and homomorphic encryption, to guarantee information security and guarding techniques, e.g., secure communication protocol and trusted execution environments for system security \cite{tang2022secure}.
Nevertheless, these privacy-preserving tools usually come at a cost of performance, which still demands careful design specific to collaborative edge training.

\section{Conclusion}
Big AI models have unlocked a wide range of intelligent services at the edge and promote the need of edge model training tasks like personalized fine-tuning and continual model refinement.
In this article, we propose collaborative edge training, a novel training mechanism distinct from traditional centralized cloud training and on-device training, which breaks the resource wall across a set of trusted edge devices for sustainable, expedited, and private edge model training.
We build a comprehensive framework to unify the workflow of collaborative edge training and analyze in-depth its merits and research questions to be addressed in each phase of its lifecycle.
As a case study, we further investigate how parallelism design impacts the energy demand of collaborative training, and explain design choices with prototype-based experiments.
We also discuss open challenges toward sustainable collaborative edge training from a full-stack perspective, suggesting future directions for sustainability research in the big AI model era. 

\bibliographystyle{IEEEtran}
\bibliography{main.bib}

\end{document}